\documentclass[sigconf]{acmart}
\AtBeginDocument{%
  \providecommand\BibTeX{{%
    \normalfont B\kern-0.5em{\scshape i\kern-0.25em b}\kern-0.8em\TeX}}}

\usepackage{multirow}
\usepackage{xcolor}
\copyrightyear{2020}
\acmYear{2020}
\setcopyright{acmcopyright}\acmConference[MM '20]{Proceedings of the 28th ACM
International Conference on Multimedia}{October 12--16, 2020}{Seattle, WA, USA}
\acmBooktitle{Proceedings of the 28th ACM International Conference on Multimedia
(MM '20), October 12--16, 2020, Seattle, WA, USA}
\acmPrice{15.00}
\acmDOI{10.1145/3394171.3413916}
\acmISBN{978-1-4503-7988-5/20/10}

\begin{document}
\fancyhead{}
\title{Interpretable Embedding for Ad-Hoc Video Search}

\author{Jiaxin Wu}
\email{jiaxin.wu@my.cityu.edu.hk}
\affiliation{%
  \institution{City University of Hong Kong}
  \city{Hong Kong}
}

\author{Chong-Wah Ngo}
\email{cscwngo@cityu.edu.hk}
\affiliation{%
  \institution{City University of Hong Kong}
  \city{Hong Kong}
}

\renewcommand{\shortauthors}{Wu and Ngo}
\newcommand{\tabincell}[2]{\begin{tabular}{@{}#1@{}}#2\end{tabular}}  
\begin{abstract}
Answering query with semantic concepts has long been the mainstream approach for video search. Until recently, its performance is surpassed by concept-free approach, which embeds queries in a joint space as videos. Nevertheless, the embedded features as well as search results are not interpretable, hindering subsequent steps in video browsing and query reformulation. This paper integrates feature embedding and concept interpretation into a  neural network for unified dual-task learning. In this way, an embedding is associated with a list of semantic concepts as an interpretation of video content. This paper empirically demonstrates that, by using either the embedding features or concepts, considerable search improvement is attainable on TRECVid benchmarked datasets. Concepts are not only effective in pruning false positive videos, but also highly complementary to concept-free search, leading to large margin of improvement compared to state-of-the-art approaches.
\end{abstract}



\maketitle

\section{Introduction}
\label{intro}
Ad-hoc video search (AVS) is to retrieve video segments for textual queries. As no visual example is provided along with the textual query, AVS is also known as zero-example video retrieval. The task is challenging due to the semantic gap between the high-level semantics expressed in terms of textual description and the low-level signals embedded in videos. Furthermore, AVS is ad-hoc for assuming the scenario of open vocabulary, where the composition of a query using words or concepts new to a search engine is allowed. In the literature, TRECVid AVS is the most known activity that conducts benchmark evaluation for this task \cite{Trecvid2016}. The evaluation is conducted on a large video collection, e.g., V3C1 corpus \cite{V3C1} composed of one million video segments, with no training data or labels being provided. The testing query is non-verbose, e.g., \textsl{Find shots of a woman wearing a red dress outside in the daytime}. Achieving satisfactory performance in face of large datasets, short query information and open vocabulary problem is certainly difficult, as evidenced from the evaluation results of TRECVid on automatic search \cite{Trecvid2016,Trecvid2018,Trecvid2019,TRECVid2017}.

The mainstream approaches are devoted to query understanding, either relying on concept classifiers (i.e., concept-based \cite{Waseda_Meisei2017,mediamill2017,ITI-CERTH2018,ontology_videosearch}) or learning embedding features common to text and video (i.e., concept-free \cite{w2vvpp,dualconding,ALT2019}) for search. Concept-based approach explicitly maps a user query to concept tokens. Capitalizing on the pre-trained convolutional neural network (CNN), a variety of concepts (e.g., object, action, scene) in videos are indexed and then matched against the tokens extracted from query description. As similarity is semantically enumerated based on their common concepts, the search result is explainable. The progress of concept-based search is bottlenecked by a number of issues, including the selection of concepts for query \cite{Waseda_Meisei2017} and the reliability of concept classifiers \cite{mediamill2017}. Concept-free approach, in contrast, bypasses these issues and performs matching between query and video by their embedded features. The features are learnt in a black box manner by minimizing the distances of video-text pairs. As features are not interpretable, the result of matching cannot be unrolled to recount the commonality between video-text pairs as in concept-based search.

The success of concept-based approach for AVS has been evidenced in \cite{mediamill2017,Waseda_Meisei2017}, by building a large visual vocabulary to accommodate for tens of thousands of semantic concepts. Screening these concepts to represent user queries, however, is not a trivial issue. Human intervention is often required. As studied in interactive search \cite{VBS}, human is excellent in picking the right concepts for video search. Concept-free approach, which is relieved from the burden of concept screening, has recently surpassed the retrieval performance of concept-based approach in automatic search \cite{Trecvid2018,Trecvid2019,w2vvpp}. Nevertheless, being not interpretable, its utility for browsing is questionable when user inspects the search result. Unlike concept-based search, there is also no explicit way to identify the mismatch of a video from the query when the performance is not satisfying.

While both concept-based and concept-free approaches have their respective merit and limitation, there is no research studying the complementarity between them. This paper proposes a new network architecture to equip concept-free approach with the capability of inferring concepts from an embedded feature. The network performs dual tasks, i.e., learning cross-modal embedding features while enforcing the features to explicitly recount the concepts underlying a video. The network is trained end-to-end to force the consistency between features and concepts. This results in each video being associated with an embedding feature for search and a concept list for interpretation. Empirically, this paper shows that, by combining both features and concepts for search, superior performance is attainable on the TRECVid benchmarked datasets. The main contribution of this paper is the proposal of a new architecture along with its novel loss function for concept decoding. This paper also paves new insights of how concept-free and concept-based approaches can possibly see eye-to-eye for a number of issues in AVS. These issues include using the decoded concepts to verify the result of concept-free search, and the ability of handling Boolean queries.

\section{Related Work}
\label{relatedwork}

Ad-hoc video search, being a long-lasting task annually evaluated in TRECVid, can date back to as early as year 2003 \cite{trecvid2006}. The task is, presented with a topic (i.e., textual descriptions of information need), a search system formulates query and then returns a ranked list of video shots. Since its beginning, the utilization of concepts for search has become the main focus of study \cite{Snoek:concept_video_retrieval}. The progress evolves from the early efforts of concept bank development and ontology reasoning \cite{Naphade:LSCOM,Jiang:semantic_concept,Snoek:challenge} to the recent advances in concept screening, representation and combination \cite{NII2016,ITI-CERTH2018,Lu:ICMR,Waseda2018,Informedia2018}. This branch of approaches is generally referred to as {\bf concept-based search}. The most recent approaches mostly focus on the linguistic analysis for concept selection on large vocabulary \cite{Waseda2018,Informedia2018}. To deal with out-of-vocabulary (OOV) problem, query expansion with ontology influencing \cite{Waseda2018} and webly-label learning by crawling online visual data \cite{Informedia2018} are commonly adopted. Despite numerous progress \cite{Boer:Semantic,Jiang:semantic_gap,vireoAVS2019}, automatic selection of concepts to capture query semantics as well as context remains highly difficult. Human intervention is often required in practice, for example, by manually removing undesired concepts after automatic matching \cite{anh:KIS} or by hand-picking query phrases that should be matched with concepts \cite{WasedaMeiseiSoftbank2019}. 

Concept-based search is superior in finding videos when the concepts required for a query topic can be precisely identified for search. The associated problem, however, is the inherent expression ambiguity when using a sparse list of concepts to describe a complex information need. For example, a topic of \textsl{Find shots of a person holding a tool and cutting something} is hard to be expressed precisely using the concepts like \lq\lq holding\rq\rq, \lq\lq cutting\rq\rq, and \lq\lq tool\rq\rq, especially if these concepts are treated independently during search. 
To this end, {\bf concept-free search}, which embeds the entire video shot and information need as high-dimensional features, is recently shown to be more effective in capturing compositional concepts \cite{w2vvpp,dualconding,Informedia2018}. Nevertheless, the performance of concept-free search is not always predictable. For example, the queries \textsl{Find shots of people queuing} and \textsl{Find shots of people standing in line}, despite being similar in meaning, can end up in considerably different retrieval performances.

With the availability of image and video captioning datasets (e.g., MSCOCO \cite{MSCOCO}, MSVD \cite{msvd}, MSR-VTT \cite{msr-vtt}), the training of visual-text embedding models is greatly facilitated. Various models have been attempted for AVS, including VideoStory \cite{videostory}, visual semantic embedding (VSE++) \cite{vse}, intra-modal and inter-modal attention networks \cite{Informedia2018}, Word2VisualVec (W2VV) \cite{w2vv} and dual encoding \cite{dualconding}. These models differ mainly in ways of how query topics are represented and encoded. For instance, VSE++ employs a recurrent network to encode the word sequence \cite{vse}, while attention networks weight the text and visual elements for embedding \cite{Informedia2018}. More sophisticated approach is W2VV \cite{w2vv}, which encodes bag-of-words, word2vec and word sequence altogether. The extension of W2VV (W2VV++) is the first model that shows significant improvement in AVS over the concept-based approaches \cite{Trecvid2018}. Building on top of W2VV++, the most recent dual encoding network \cite{dualconding} reports state-of-the-art performance by multi-level encoding. Three different levels of information are considered, including the short-term pooling of local features and mean pooling of word and sequence features respectively. Different from other models, the network treats a video frame as a word and processes video shot in a similar fashion as a text sequence. Similar approach is adopted in \cite{ALT2019}, with the use of graph convolutional network and VLAD for encoding.

A hybrid of concept-based and concept-free approaches has also been studied  \cite{vireo2017,Informedia2018,Waseda2018,WasedaMeiseiSoftbank2019,EURECOM2019,Kindai_kobe2019,mediamill2017}. The early works include the fusion of VideoStory embedding and concept features \cite{mediamill2017}, which leads to improvement over individual features. In the recent TRECVid benchmarking, late fusion of concept-based and concept-free approaches has become a norm \cite{vireo2017,Informedia2018,Waseda2018,WasedaMeiseiSoftbank2019,EURECOM2019,Kindai_kobe2019}. Although being shown to be complementary, the fact that both features are generated using two separate models trained using different forms of data has tremendously increased the system complexity. Different from these works, the proposed dual-task model is end-to-end trained using the same set of data for generating both embedding and concept features. This results in significant save of effort in training and housekeeping models, while maintaining the consistency between two different natures of features.

\section{Dual-Task Network}

\label{method}
 \begin{figure*}[h] 
        \centering
        \includegraphics[width=0.85\textwidth]{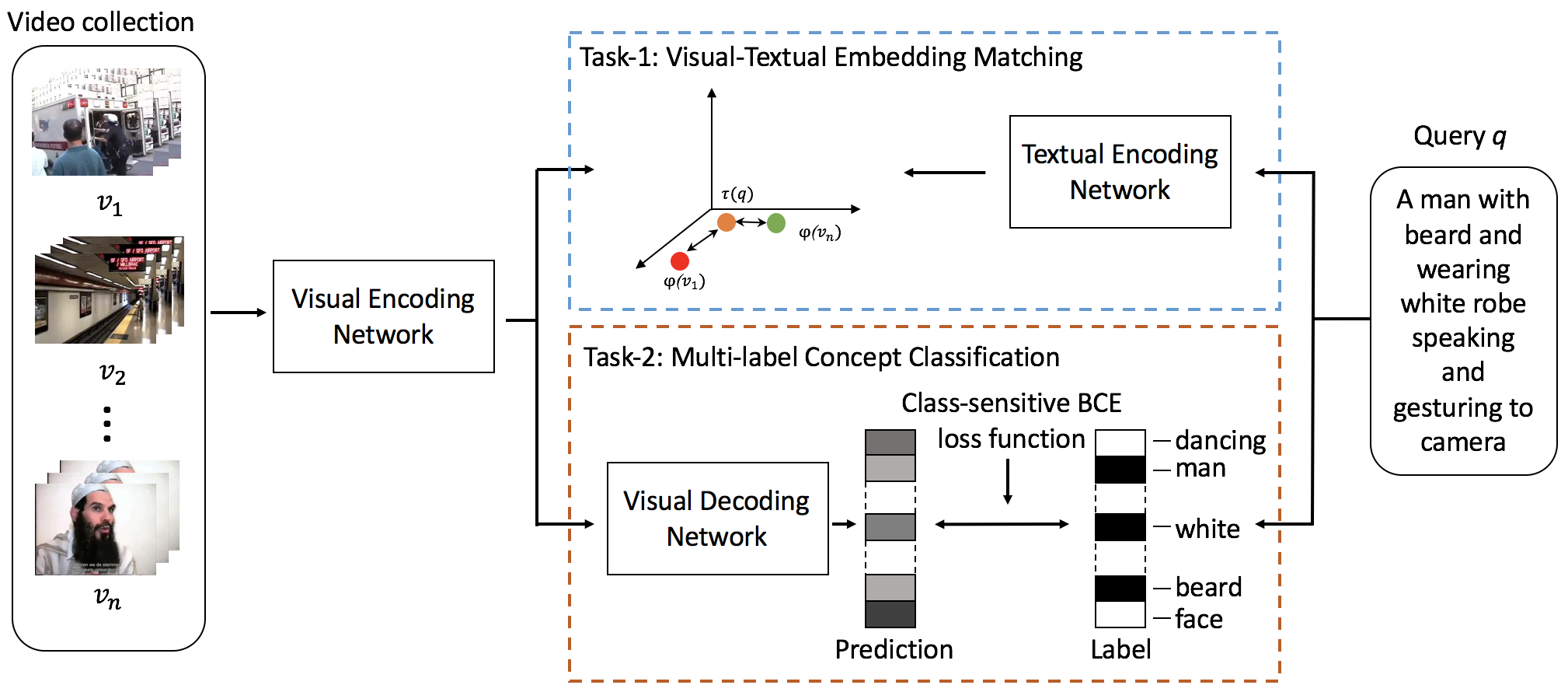}
        \caption{An overview of the end-to-end dual-task network architecture}
        \label{fig:archicture} 
 \end{figure*}  
Figure \ref{fig:archicture} illustrates the architecture of dual-task network, which is composed of visual-textual embedding (Task-1) and multi-label concept classification (Task-2). Both tasks share visual encoding network for video feature extraction. Taking user query as input, Task-1 trains a textual encoding network to project the query into the same latent space as video features. Task-2, on the other hand, trains a visual decoding network to predict the concepts underlying a video feature. Given a video-text pair, Task-1 aims to minimize the pairwise distance, while Task-2 is to recover the query words from the video embedding. It is worth noting that a query usually describes only one perspective of video content. Hence, the loss function of decoder is novelly designed to emphasize more on missing query words than the words outside of a query. To this end, as both tasks are learnt simultaneously, the video features are learnt iteratively to accommodate both tasks.

\subsection{Visual-text Embedding}
We use the same architecture as dual encoding network \cite{dualconding} to extract multi-level video and text features. The visual encoding network has three sub-networks. The first sub-network captures the global information of a video segment by mean pooling its CNN features over the video frames. The second sub-network extracts the temporal information of video by mean pooling the features of the bidirectional GRU \cite{gru} (biGRU) across all time steps. The final sub-network further extracts the local information in the temporal sequence of video by performing 1-d CNN on the top of the biGRU features from the second sub-network. A multi-level video feature is formed by concatenating the global, temporal and local information extracted from these sub-networks. The visual embedding is denoted as 
\begin{equation}
f(v)=[f_{v}^{(1)},f_{v}^{(2)},f_{v}^{(3)}]
\end{equation}
where $f_{v}^{(i)}\ (i=1,2,3)$ denotes the feature extracted from the first, second and third sub-networks for a video $v$. The video feature $f(v)$ is further mapped to the common space of $d$ dimensions by using a fully connected layer (FC) and a batch normalization layer (BN) as following
\begin{equation}
\phi(v) = BN(\mathbf{W}_vf(v)+b_v),\phi(v)\in\mathbb{R}^d
\end{equation}
where BN($\cdot$) denotes the batch normalized layer. The matrix $\mathbf{W}_v$ and the vector $b_v$ are weight and bias of the fully connected layer in the visual encoding network. 

The query $q$ is encoded similarly using the texture encoding network. Each word in $q$ is represented as a one-hot vector $m_i$ and then concatenated as a matrix $\mathbf{M}_q=\{m_1,m_2,...,m_i,...,m_n\}$ to represent the sentence $q$. The first level feature $f_{q}^{(1)}$ is obtained by taking the mean value of the matrix $\mathbf{M}_q$ over all the vectors of sentence words. A denser vector is further generated for every word by multiplying its one-hot vector with a pre-trained Word2Vec matrix. The second-level feature of the query $f_{q}^{(2)}$ is obtained by mean pooling of the features output by each stage of biGRU \cite{gru} over the sequence of dense vectors. The third-level feature $f_{q}^{(3)}$ is extracted by conducting 1-d CNN over the biGRU features obtained from the second level. Finally, a multi-level feature for the query $q$ is formed by concatenating all of the features as

\begin{equation}
f(q)=[f_{q}^{(1)},f_{q}^{(2)},f_{q}^{(3)}]
\end{equation}
The feature is subsequently mapped to the common space as the video feature by a fully connected layer and a batch normalize layer
\begin{equation}
\tau(q) = BN(\mathbf{W}_qf(q)+b_q), \tau(q)\in\mathbb{R}^d
\end{equation}
where $\mathbf{W}_q$ and $b_q$ are the weight and bias of the fully connected layer in the textual encoding network. The textual embedding has the same output dimension $d$ as visual embedding.

Finally, the similarity of the input query $q$ and the video $v$ is computed as $S(v,q) = sim(\phi(v),\tau(q))$. The visual-textual embedding matching task is trained in an end-to-end manner by using improved marginal ranking loss \cite{vse}. The loss in this matching task is defined as
\begin{equation}
\begin{aligned}
\label{loss_matching}
loss_{matching}(v,q) = max(0,c+S(v^{\_},q)-S(v,q)) \\
+max(0,c+S(v,q^{\_})-S(v,q))
\end{aligned}
\end{equation}
where $c$ is the margin. The visual-text pair is indicated by the notations $v$ and $q$ respectively. The query $q$ is associated with a negative sample $v-$, and similarly $v$ is associated with a negative query $q-$. The visual-textual embedding matching task is trained to minimize this loss.  

\subsection{Multi-label Concept Decoding}

Video content is multi-perspective. The content can be narrated from different perspectives depending on the subjects or contexts of interest in the topic of discussion. Even for one perspective, there exists a variety of ways expressing the same semantics using different words. Hence, given a video, Task-2 is to train a decoder such that all the words in its descriptions are activated with high probability. The activated words that are absent from the descriptions do not necessarily belong to false positives. Instead, these words may be just not mentioned than being classified incorrectly. In general, the challenge of training such a decoder is beyond concept prediction as in the conventional CNN. The novelty of Task-2 is the design of a loss function to maximize the intersection between the classified concepts of a video embedding and the words mentioned in the captions.  

The visual decoding network is comprised of a fully connected layer and a batch normalization layer. The output is denoted as
\begin{equation}
g(v) = BN(\mathbf{W}_d\phi(v)+b_d)
\end{equation}
where $\mathbf{W}_d$ and $b_d$ are weight and bias of the network. The sigmoid function is performed over the output to obtain the probability distribution $\hat{y}_v$ for a video embedding $g(v)$
\begin{equation}
\hat{y}_v = sigmoid(g(v)).
\end{equation}
where $\hat{y}_v=[\hat{y}_{v1},\hat{y}_{v2},...,\hat{y}_{vi},...,\hat{y}_{vm}]  \ (\hat{y}_{vi}\in \mathbb{R}^+)$ for $m=11,147$ numbers of concept classes. The value of $m$ is compiled from the training set, by including the words that appear in at least five descriptions and removing the words in the NLTK stopword list. For a video $v$, the words that appear in its captions are labeled as ground-truth $y_v=[y_{v1},y_{v2},...,y_{vi},...,y_{vm}]$, where $y_{vi}\in\{0,1\}$ indicates whether a word is present in the captions of $v$. 

Binary cross entropy (BCE) loss is widely adopted for multi-label classification task. However, BCE cannot be directly applied for treating each class equally by penalizing the predictions different from the ground-truth labels. In general, the number of words used to describe a video is much smaller than the total number of concepts. Using BCE will cause the amount of loss being dominated by the words not mentioned in video captions. The decoder trained in this way will play trick to predict almost all classes as zero in order to get an overall low BCE loss. Thus, a class-sensitive loss is proposed, by computing the loss of mentioned words separately from those that do not appear in the captions, as following
\begin{equation}
\begin{aligned}
\label{eq:newlossfunction}
loss_{classification}(v,y_v) = \lambda \frac{1}{\sum_i^m y_{vi}}\sum_i^m y_{vi}BCEloss_i \\+(1-\lambda) \frac{1}{\sum_i^m (1-y_{vi})}\sum_i^m (1-y_{vi})BCEloss_i,
\end{aligned}
\end{equation}

\begin{equation}
\label{eq:BCEloss}
BCEloss_i= -[y_{vi}log(\hat{y}_{vi})+ (1-y_{vi})log(1-\hat{y}_{vi})].
\end{equation}
Eq. (\ref{eq:BCEloss}) is the BCE loss function that computes the binary cross-entropy for a class $i$. The proposed function in Eq. (\ref{eq:newlossfunction}) is the summation of two BCE terms. The first term computes the losses of those classes mentioned in the captions, i.e., $y_{vi}=1$ in the ground truth of a video. The second term computes the losses of the remaining classes, i.e., $y_{vi}=0$. Note that the proportion of concepts involved in the first and second terms is highly imbalanced. Eq. (\ref{eq:newlossfunction}) effectively assigns higher weights to the mentioned concepts. In this way, the decoder is trained to play a more active role in maximizing the overlap between the predicted and ground-truth concepts. Nevertheless, to avoid an excessive number of concepts being predicted as positive, a hyper parameter $\lambda \in [0,1]$ is required to keep a balance between activating mentioned words and suppressing irrelevant concepts. When $\lambda$ equals to 1 or 0, it represents the extreme case of whether to penalize only the concepts being mentioned (i.e., $\lambda=1$) or vice versa (i.e., $\lambda=0$). In practice, the value of $\lambda$ should bias towards the second term to restrict the number of activated concepts. This is also justified in the experiment, where the value is learnt as $\lambda=0.2$, when verified using validation set. The multi-class classification is trained in an end-to-end manner to minimize the $loss_{classification}$.

\subsection{Dual-task Learning and Inference}
We have designed one loss for each task and they are $loss_{matching}$ and $loss_{classification}$. The dual-task loss is set as
\begin{equation}
\label{eq:combinedloss}
loss_{combined}= loss_{matching}+ loss_{classification}.
\end{equation}
In the training stage, we compute the average dual-task loss over all input video-text pairs in a batch and update all the parameters of the model simultaneously by using this loss.

In the inference stage, we have two trained models for video search. One model is from the visual-textual embedding matching task (denoted as $DT_{embedding}$) and the other model comes from the multi-label video concept classification task (denoted as $DT_{concept}$). The embedding model $DT_{embedding}$ and the concept model $DT_{concept}$ have the same parameters in the visual encoding network. For all videos in the collection, we first use the trained visual encoding network to extract their visual embeddings and apply the trained visual decoding network of $DT_{concept}$ to get their predicted concepts offline. As a result, each video $v$ in the dataset is indexed with an embedding $\phi(v) \in \mathbb{R}^d$ and a predicted concept vector $\hat{y}_v\in \mathbb{R}^{m+}$. Given a test query, the embedding model $DT_{embedding}$ applies its textual encoding network to encode the test query as a textual embedding $\tau(q) \in \mathbb{R}^d$. Then, $DT_{embedding}$ will give each video $v$ a score ($score(v,q)_{embedding}$) according to the similarity of its embedding with the textual embedding of the query $q$
\begin{equation}
\label{eq:score_emebdding}
score(v,q)_{embedding}= sim(\phi(v),\tau(q)).
\end{equation}
$DT_{embedding}$ will output a list of videos for the query ranked according to the scores. The top-1 in the list has the highest score which represents the video best matches the input query. Meanwhile, the concept model $DT_{concept}$ will generate a one-hot vector for the given test query. It maps the query sentence $q$ over the constructed concept list from the classification task to get a one-hot query vector $c_q = [c_{q1},c_{q2},...,c_{qi},...c_{qm}],c_{qi}\in\{1,0\}$. If $c_{qi}=1$, it means that $i$th concept of the concept list is wanted by the query. Then, $DT_{concept}$ will also give each video $v$ a score ($score(v,q)_{concept}) $ based on the similarity of its predicted concept vector $\hat{y}_v$ and the query vector $c_q$. 
\begin{equation}
\label{eq:score_concept}
score(v,q)_{concept}= sim(\hat{y}_v,c_q).
\end{equation}
Another ranked list is output by the concept model $DT_{concept}$ based on this score. A linear function is used to combine two models together. We give a combined score ($score(v,q)_{combined}$) for each video $v$ with respect to the input query $q$ as
\begin{equation}
\begin{aligned}
\label{eq:score_combined}
score(v,q)_{combined} = (1-\theta)(score(v,q)_{embedding}) \\+ \theta(score(v,q)_{concept}).
\end{aligned}
\end{equation}
where $\theta$ is a hyper parameter to control the contributions of the embedding model $DT_{emebdding}$ and the concept model $DT_{concept}$ to the final retrieval score. We denote the model which uses the $score(v,q)_{combined}$ for video search as $DT_{combined}$. Normally, we use the combined model $DT_{combined}$ to retrieve videos for an input query in the experiments.

\section{Experiments}
\label{experiments}

\label{experiments}
We begin by describing the datasets, followed by ablation studies to justify the proposed class-sensitive loss function and parameter choices. Performance comparison with state-of-the-art techniques is then presented. Finally, we provide insights on the potential of dual-task model in interpreting search result and answering Boolean queries.

\subsection{Experimental Settings}
$\mathbf{Datasets}$. Table \ref{exp:datasets} lists the datasets used in the experiments. The proposed model is trained and validated on the captioning datasets. These datasets are also used by other approaches such as W2VV++ \cite{w2vvpp} and dual coding \cite{dualconding}. The number of captions per video varies from 2 in TV2016TRAIN \cite{Trecvid2016} to as many as 40 in MSVD \cite{msvd}. The performance of AVS is evaluated on two large video collections. Both are TRECVid benchmarked datasets, where IACC.3 \cite{Trecvid2016} is used during the years 2016 to 2018 and V3C1\cite{V3C1} in the year 2019. As TRECVid evaluates 30 query topics per year, a total of 120 queries are involved in the evaluation. For convenience, we name the query sets as tv16, tv17, tv18 and tv19, and each set contains the 30 queries being evaluated on that year. The full set of queries is listed in the supplementary document.
\begin{table}[t]
\caption{Datasets information.}
\label{exp:datasets}
\begin{tabular}{lccc}
    \toprule
name         & \#video   & \#caption & \#AVS test query \\
\hline
\multicolumn{4}{l}{Training set:}             \\
MSR-VTT \cite{msr-vtt}      & 10,000         & 200,000      &             \\
TGIF \cite{tgif}         & 100,855        & 124,534     &             \\
\hline
\multicolumn{4}{l}{Validation set:}             \\
TV2016TRAIN \cite{Trecvid2016}  & 200            & 400         &             \\
\hline
\multicolumn{4}{l}{Classification test set:} \\
MSVD \cite{msvd}          & 1,907           & 80,837      &             \\
\hline
\multicolumn{4}{l}{AVS test set:} \\
IACC.3 \cite{Trecvid2016}        & 335,944        &             & 90          \\
V3C1 \cite{V3C1}         & 1,082,659      &             & 30         \\
\bottomrule
\end{tabular}
\end{table}

$\mathbf{Evaluation \ metric}$. We use extended inferred average precision (xinfAP) to evaluate ad-hoc video search results \cite{Trecvid2019}. For the overall performance of a search system, we report the mean xinfAP of all test queries as results.

$\mathbf{Implementation \ details}$. 
We use the PyTorch code provided by the dual coding model \cite{dualconding} to set up the basic architecture of visual encoding network and textual encoding network. Following \cite{dualconding}, we set the output dimension of both networks as $d=2,048$, and the margin in the triplet loss function (Eq. (\ref{loss_matching})) as $c=0.2$. We use the pre-trained ResNet-152 \cite{DL_2017} and ResNext-101 \cite{mediamill2017} to extract a feature of 4,096 dimensions for each video frame. For the visual decoding network, the concept list is composed of $m=11,147$ words, compiling from the captions of TGIF and MSR-VTT. We use a learning rate of 0.0001 and Adam optimizer to train the model. The batch size is 128. The hyper-parameter is set as $\lambda=0.2$ (Eq. (\ref{eq:newlossfunction})) based on validation set, and we report $\theta=0.3$ (Eq. (\ref{eq:score_combined})) as our combined model.

\subsection{Ablation Studies}

\textbf{BCE loss}. We first compare the proposed class-sensitive loss (Eq. (\ref{eq:newlossfunction})) to the normal BCE in the design of multi-label concept decoder. The evaluation is conducted by counting the number of classified concepts that are mentioned in video captions. On MSVD dataset \cite{msvd} with 1,907 video clips, the proposed BCE loss attains recall@10=0.532. In other words, on average there are more than 50\% of concepts in the top-10 list are mentioned by the video captions. This performance is significantly better than the normal BCE loss which only manages to reach recall@10=0.293.

\begin{figure}[t]
  \centering
  \includegraphics[width=0.9\linewidth]{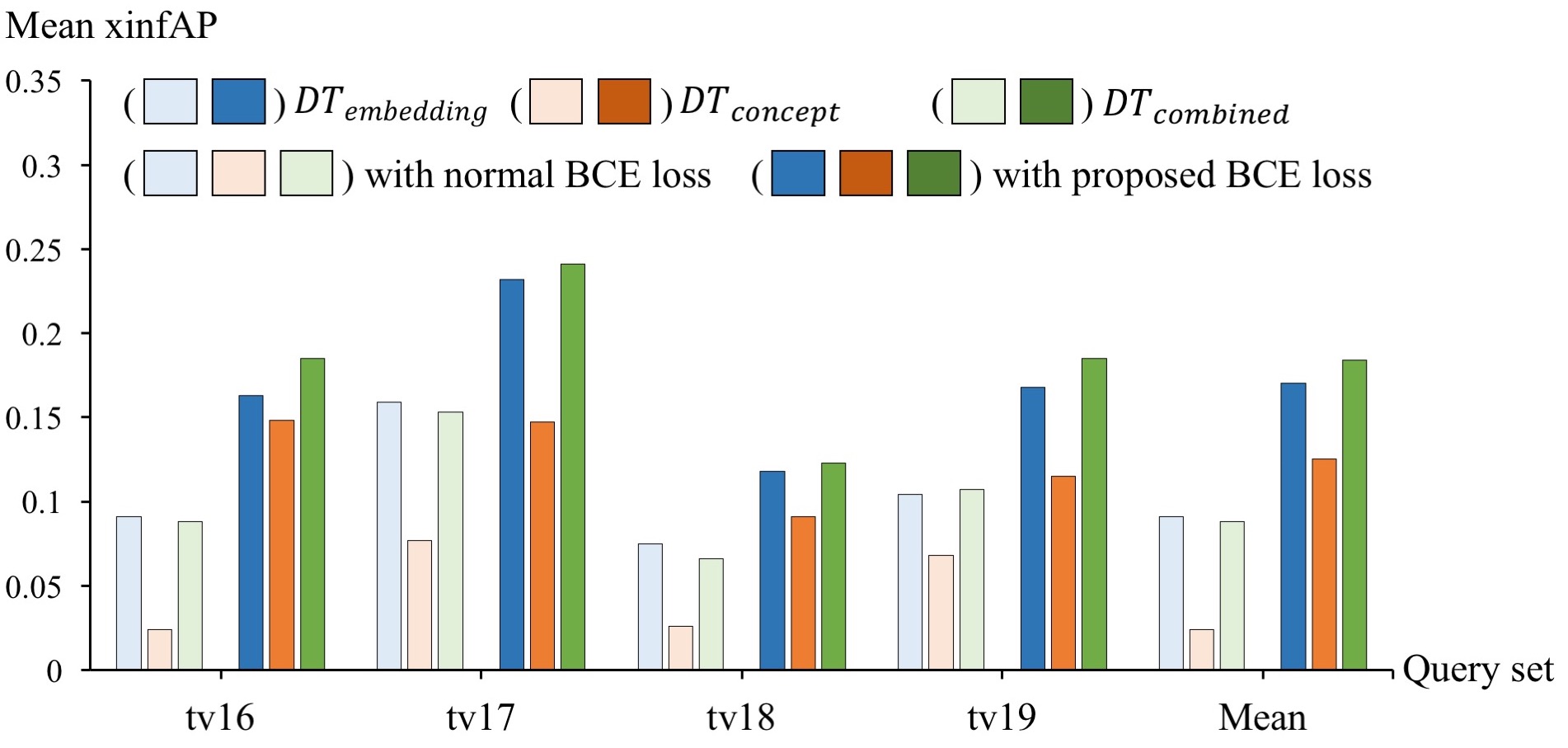}
  \caption{The AVS performance comparison between the normal BCE loss and our proposed BCE loss. }
  \label{fig:loss_comparison}
\end{figure}

Figure \ref{fig:loss_comparison} further contrasts the retrieval performances between the proposed class-sensitive and normal loss functions on four sets of TRECVid queries. Note that in the figure, two groups of performances are shown for each query set. Each group separately lists three performances for using embedding feature only (i.e., $DT_{embedding}$), concept only (i.e., $DT_{concept}$) and late fusion of them (i.e., $DT_{combined}$). The performances based on the models trained with normal and proposed BCE losses are visualized with bars of lighter and darker colors respectively. As shown in the figure, the models trained with the proposed class-sensitive loss function show consistently better retrieval rate across all the four query sets. For concept-only retrieval, the average performance over 120 queries is five times higher than the model trained with normal BCE. We attribute the performance gain due to higher accuracy in concept prediction. The result indicates the merit of the new BCE loss function for dual-task learning. With class-sensitive loss, the performance of $DT_{embedding}$ also improves. The performance difference is almost double (xinfAP =0.170) when comparing to the model trained by the normal BCE (xinfAP = 0.091). Furthermore, by using the proposed BCE loss, the late fusion of concept-only and embedding-only features always leads to performance boost across the four query sets. In contrast, the model trained with normal BCE only manages to show improvement on tv19.

\begin{table}[]
\caption{Comparison between dual-task and two single-task models}
\label{exp:dual_vs_single}
\begin{tabular}{l|ccc|c|c}
    \toprule
                       Datasets                      & \multicolumn{3}{c|}{IACC.3} & V3C1 &   \\
                       \hline
                       Query sets               & tv16  & tv17  & tv18  & tv19  & Mean  \\
                       \hline
                       Single-task models: &       &       &       &       &       \\
                       $ST_{concept}$             & 0.134 & 0.137 & 0.068 & 0.104 & 0.111 \\
                      $ST_{embedding}$                 & 0.156 & 0.222 & 0.115 & 0.160 & 0.163 \\
                       \hline
                       Dual-task models:       &       &       &       &       &       \\
                       $DT_{concept}$               & 0.148 & 0.147 & 0.091 & 0.115 & 0.125 \\
                       $DT_{embedding}$                 & \textbf{0.163} & \textbf{0.232} & \textbf{0.118} & \textbf{0.168} & \textbf{0.170} \\
  \bottomrule
\end{tabular}
\end{table}

\textbf{Dual-task learning}. Next, we verify the merit of dual-task learning versus training two single tasks independently. Specifically, the encoders for Task-1 (visual-text embedding) and the encoder-decoder for Task-2 (multi-level concept classification) are trained separately. For notation convenience, we call the two single-task models as $ST_{concept}$ and $ST_{embedding}$ respectively. As shown in Table \ref{exp:dual_vs_single}, consistent improvement is noticeable for dual-task model when performing retrieval using either concept-only or embedding-only feature. $DT_{embedding}$ improves 69 out of 120 queries over $ST_{embedding}$. Using query-556 \textsl{Find shots of a person wearing a blue shirt} as an example, xinfAP is significantly boosted from 0.144 to 0.198. Benefited from dual-task learning, true positives are ranked at higher positions in $DT_{embedding}$. At xinfAP@100, the average improvement of $DT_{embedding}$ over $ST_{embedding}$ is 18.3\%. Similar trend of improvement is also observed in $DT_{concept}$, where xinfAP is boosted for 66 out of 120 queries. Benefited from $DT_{embedding}$, some true positives originally out of the search depth of 1,000 are ranked high. Examples include query-577 \textsl{Find shots of two or more people wearing coats}, where $DT_{concept}$ brings forward more than 100 positive videos. As a consequence, xinfAP is boosted for 173\% from 0.1793 to 0.4898 for query-577. Nevertheless, there are 19 queries drop for both $DT_{embedding}$ and $DT_{concept}$. These are the queries where either $ST_{concept}$ or $ST_{embedding}$ performs poorly. As training is conducted using video captioning datasets, dual-task learning may have been confused if the results of both tasks are in conflict with each other. 

\begin{figure}[h]
  \centering
  \includegraphics[width=\linewidth]{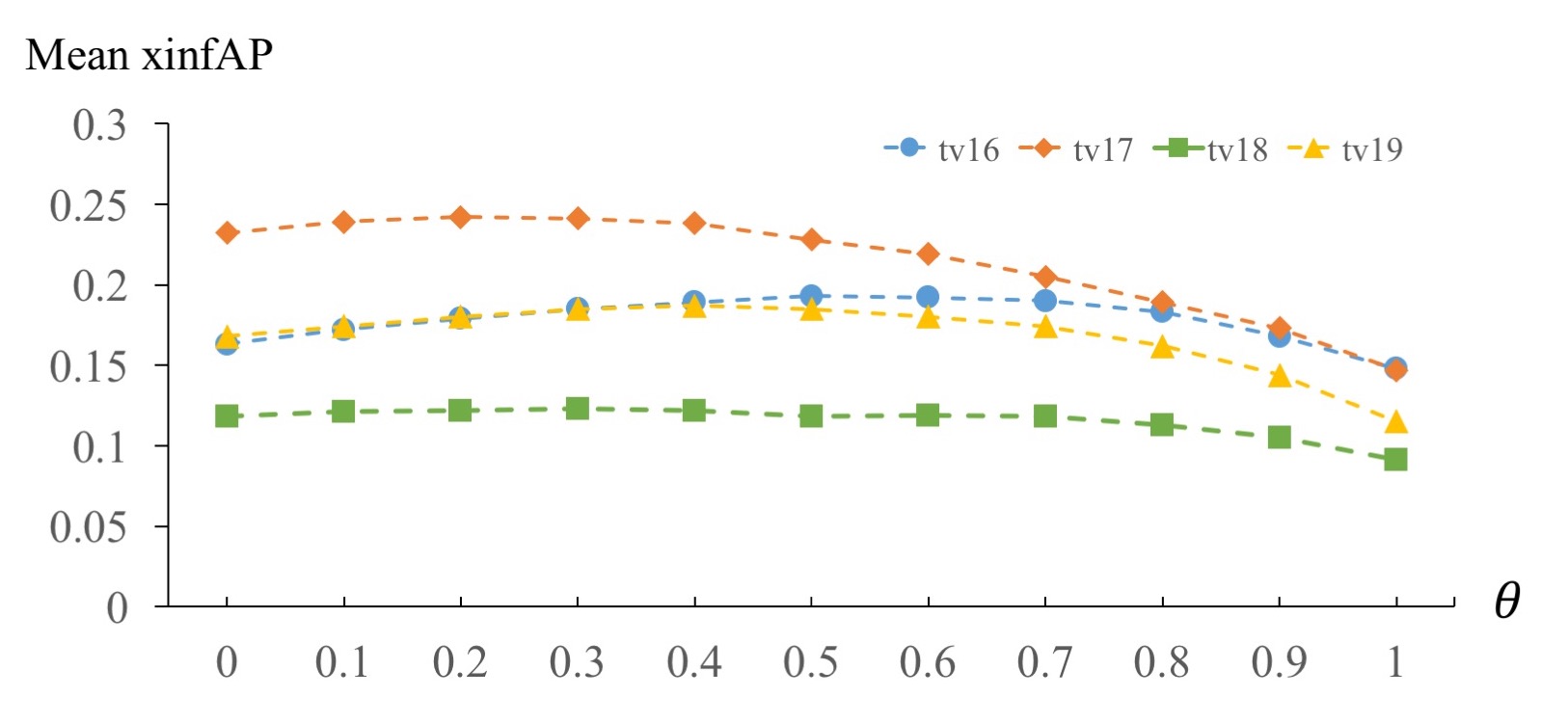}
  \caption{Sensitivity of hyper-parameter $\theta$ in the late fusion of embedding-only and concept-only searches.}
  \label{fig:diff_theta}
\end{figure}

\textbf{Late fusion}. In general, concept-only search performs better when a query topic can be uniquely described by a few concepts independently. In contrast, embedding-only search is superior for complex queries, such as query-543 \textsl{Find shots of a person communicating using sign language}, where collective understanding of query terms is required. As both search methodologies are complementary, late fusion is employed. Figure \ref{fig:diff_theta} shows the sensitivity of hyper-parameter $\theta$ (Eq. (\ref{eq:score_combined})) that linearly combines the rank lists of both methods. The extreme values represents retrieval by embedding-only (i.e., $\theta=0$) and concept-only (i.e., $\theta = 1.0$). As observed, the retrieval performance varies slightly when $\theta$ is set in the range of $[0.2,0.4]$. Considering the capability of answering complex queries, we set $\theta=0.3$ to bias embedding-only more than concept-only retrieval in the late fusion. The detailed performances of our models on every query are list in the supplementary material.

\subsection{AVS Performance Comparison}
\label{sec_avs_comparison}
\begin{figure*}[t] 
  \centering
  \includegraphics[width=\linewidth]{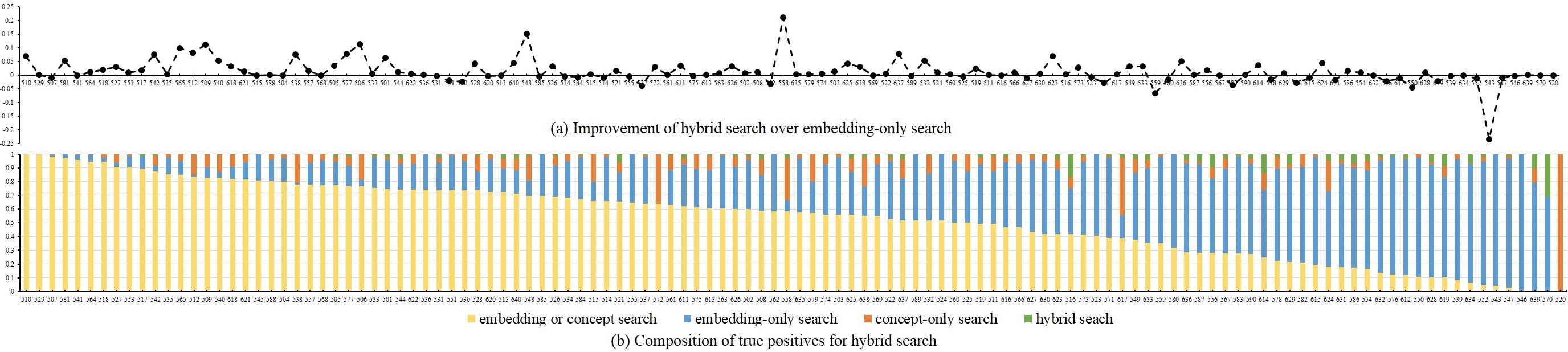}
  \caption{Visualization showing (a) the improvement of $DT_{combined}$ over $DT_{embedding}$ and (b) composition of true positives in $DT_{combined}$ within the search depth of 1,000. The x-axis shows the query topic ID. Each column in (b) visualizes the true positives that a method can retrieve with a different color.} 
  \label{fig:composition}
\end{figure*} 

We compare dual-task model to the existing methods based on embedding-only, concept-only and hybrid search. Embedding-only search includes W2VV++ \cite{w2vvpp} and dual coding \cite{dualconding}, which report the state-of-the-art performance on TRECVid datasets. Concept-based search includes mainly \cite{Markatopoulou:AVS,Waseda2016,Waseda_Meisei2017,Waseda2018,WasedaMeiseiSoftbank2019}. QKR (query and keyframe representation) \cite{Markatopoulou:AVS} performs linguistic analysis to select concepts for queries. Through word2vec \cite{word2vec}, the concept lists of queries and keyframes are projected into an embedding space for similarity comparison. Large concept bank (ConBank) composed of about 55,000 words is leveraged in \cite{Waseda_Meisei2017,Waseda2018,WasedaMeiseiSoftbank2019}. Multi-level query analysis from words, phrases to sentences is conducted for concept mapping. As this process is error-prone, manual selection of query words and phrases for mapping with concepts is also performed in \cite{Waseda2016,Waseda_Meisei2017,Waseda2018,WasedaMeiseiSoftbank2019} to contrast the performances between automatic and manual searches. The performances of hybrid approaches are mostly reported in TRECVid benchmarking only. For reference, the top-3 performances benchmarked in TRECVid are listed for comparison.

\begin{table}[b]
\caption{Comparison with other approaches. The results are cited directly from the papers. The symbol \lq /\rq\ indicates the result is not reported. Results based on hybrid search are marked with \textbf{asterisk}.} 
\label{exp:comparison}
\resizebox{\linewidth}{!}{
\begin{tabular}{l|ccc|c}
    \toprule
Datasets                     & \multicolumn{3}{c|}{IACC.3}               & V3C1    \\
\hline
Query sets                         & tv16           & tv17           & tv18           & tv19           \\
                          \hline
\multicolumn{5}{l}{TRECVid top results:}                                                  \\
Rank 1     & 0.054 \cite{NII2016}  & 0.206* \cite{mediamill2017} & 0.121 \cite{w2vvpp} & 0.163 \cite{ALT2019}         \\
Rank 2     & 0.051 \cite{ITI2016}  & 0.159 \cite{Waseda_Meisei2017}  & 0.087* \cite{Informedia2018} & 0.160 \cite{RUCMM2019}           \\
Rank 3     & 0.040 \cite{Informedia2016}   & 0.120* \cite{vireo2017}  & 0.082 \cite{NTU_rose2018}         & 0.123 \cite{WasedaMeiseiSoftbank2019}          \\
\hline
\multicolumn{5}{l}{Embedding only:} 
\\
VideoStory \cite{videostory} & 0.087          & 0.150           & /              & /              \\
VSE++ \cite{vse}           & 0.123          & 0.154          & 0.074          & /              \\
W2VV \cite{w2vv}                      & 0.050           & 0.081          & 0.013          & /              \\
W2VV++ \cite{w2vvpp}       & 0.163          & 0.196           & 0.115          & 0.127          \\
Dual coding \cite{dualconding}              & 0.165          & 0.228           & 0.117         & 0.152      \\ 
\hline
\multicolumn{5}{l}{Concept only:} \\
QKR \cite{Markatopoulou:AVS}  & 0.064          & /              & /              & /              \\
ConBank (auto)  & /  & 0.159 \cite{Waseda_Meisei2017}  & 0.060 \cite{Waseda2018}        & /              \\
ConBank (manual)   & 0.177 \cite{Waseda2016}   & 0.216 \cite{Waseda_Meisei2017}  & 0.106 \cite{Waseda2018}        & 0.114 \cite{WasedaMeiseiSoftbank2019}           \\

\hline
\multicolumn{5}{l}{Dual-task:}                                                      \\
$DT_{concept}$            & 0.148            & 0.147           & 0.091         & 0.115           \\
$DT_{embedding}$            & 0.163           & 0.232          & 0.118          & 0.168           \\
$DT_{combined}$             & \textbf{0.185*} & \textbf{0.241*} & \textbf{0.123*} & \textbf{0.185*}  \\
\bottomrule
\end{tabular}
}
\end{table}

Table \ref{exp:comparison} lists the performance comparison. Except on tv16, the proposed $DT_{embedding}$ shows better performances than its counterparts including W2VV++ and dual coding. Among the embedding-only approaches, $DT_{embedding}$ shows the highest xinfAP in 43 out of 120 queries. Through dual-task training, $DT_{embedding}$ shows higher ability in answering queries with unique concepts such as \lq\lq scarf\rq\rq\ in query-558 and \lq\lq coats\rq\rq\ in query-577. For concept-only search, $DT_{concept}$ outperforms most of its counterparts including QKR. Despite that the number of concepts is almost five times smaller than the large concept bank \cite{Waseda_Meisei2017,Waseda2018,WasedaMeiseiSoftbank2019}, $DT_{concept} $ still shows competitive performance. Nevertheless, $DT_{concept}$ is not competent with \cite{Waseda2016,Waseda_Meisei2017,Waseda2018} when human intervention is allowed for concept selection. $DT_{combined}$, as a hybrid model, outperforms all the existing approaches, including the top performers benchmarked in TRECVid. The performance also surpasses \cite{Waseda2016,Waseda_Meisei2017,Waseda2018,WasedaMeiseiSoftbank2019} which is based on manual search. Compared to the best reported result by dual coding \cite{dualconding}, the xinfAP gain is as large as 12.1\% and 21.7\% on tv16 and tv19 respectively. To verify the performance is not by chance, we conduct randomization test \cite{randomization_test}. At the $p$-value $\leq $ 0.01, the test shows that the dual-task model is significantly better than other approaches including dual coding, W2VV++ and the best reported performance in TRECVid. 

In TRECVid \cite{Trecvid2016}, the 120 query topics are classified into 12 different types of complexities based on the composition of object, scene, action and location. Comparing to dual coding, $DT_{embedding}$ is better in answering person-related queries, especially, the combination of person, object and location. The performance of $DT_{embedding}$ exceeds dual coding on all queries in this kind, e.g., query-624 \textsl{Find shots of a person in front of a curtain indoor}. Compared with dual coding and W2VV++, $DT_{concept}$ has higher xinfAP on detecting object-only queries. For some queries such as query-512 \textsl{Find shots of palm trees}, xinfAP is as much as 167.8\% better than W2VV++ and 82.7\% than dual coding. The hybrid model, $DT_{combined}$ outperforms dual coding in 76 out of 120 queries, and the xinfAP gain ranges from 0.3\% to 158.4\%. These queries correspond to person-related or object-related queries, e.g., 121.8\% gain on query-509 \textsl{Find shots of a crowd demonstrating in a city street at night}, and 100.0\% gain on query-572 \textsl{Find shots of two or more cats both visible simultaneously}. The remaining queries suffering from xinfAP degradation mostly are those queries that $DT_{concept}$ fails miserably. The drop ranges from 0.2\% to 87.5\%. For example, dual coding has a xinfAP of 0.194 on query-514 \textsl{Find shots of soldiers performing training or other military maneuvers}, while $DT_{combined}$ only manages to attain xinfAP = 0.181. For this particular query, the performance gap between $DT_{embedding}$ (xinfAP=0.191) and $DT_{concept}$ (xinfAP=0.079) is relative large. Further late fusion hurts the overall retrieval rate.

\begin{table*}[t]
\caption{Boolean query answering: comparison of dual-task model with dual coding \cite{dualconding} and concept search \cite{anh:KIS}.}
\label{exp:bool}
\begin{tabular}{|c|c|c|c|c|c|c|}
\hline
\multirow{3}{*}{Type} & \multicolumn{4}{c|}{Concept-free}                               & \multicolumn{2}{c|}{Concept-based} \\ \cline{2-7} 
                      & \multicolumn{2}{c|}{Single query} & \multicolumn{4}{c|}{Multiple sub-queries}                        \\ \cline{2-7} 
                      & Dual coding     & $DT\_{embeding}$    & Dual coding & $DT\_{embedding}$ & Nguyen et al. \cite{anh:KIS} & $DT\_{concept}$ \\ \hline
AND                   & 0.249           & 0.260           & 0.073       & 0.089         & 0.293                & 0.454       \\ \hline
OR                    & 0.714           & 0.686           & 0.737       & 0.775         & 0.356                & 0.864       \\ \hline
NOT                   & 0.045           & 0.049           & 0.353       & 0.406         & 0.067                & 0.133       \\ \hline
\end{tabular}
\end{table*}

To provide further insights, Figure \ref{fig:composition} visualizes the composition of true positives in the rank lists of $DT_{combined}$. As observed, 72 out of 120 queries share more than 50\% of true positives common between $DT_{embedding}$ and $DT_{concept}$. Among them, 50 queries show improvement in xinfAP due to elevation of ranking positions for true positions after score combination. Overall, performance improvement is more apparent for queries where true positives are separately contributed by embedding-only and concept-only searches. Finally, $DT_{combined}$ manages to pull true positives outside of 1,000 search depth from both $DT_{embedding}$ and $DT_{concept}$ for 76 queries, and the number ranges from 1 to 20. However, as these positives are still ranked low in the list, their contributions to overall xinfAP are not significant. 

\subsection{Interpretability of Concept-free Search }
\label{interpretable}
Concept-free search suffers from the black-box retrieval of results. With dual-task model, nevertheless, interpretation of search result is feasible by listing out the decoded concepts alongside the retrieved videos. Here, we conduct an experiment to verify the top-10 retrieved videos of $DT_{embedding}$ by the decoded concepts. A video is pruned from search list if the required keywords are not present among the list of decoded concepts. For example, the required keyword for query-625 \textsl{Find shots of a person wearing a backpack} is \lq\lq backpack\rq\rq. Any retrieved videos without the \lq\lq backpack\rq\rq\ concept will be eliminated from the search result. 

The experiment is conducted as follows. A total of 30 queries, where $DT_{concept}$ exhibits higher xinfAP values, are selected for experiment. The top-10 retrieval accuracy averaged over queries is 82\%. Among these query topics, 1-3 keywords are manually specified as the required concepts to present in a retrieved video. Any videos without all the required keywords present within the list of top-30 decoded concepts will be marked as negative videos. The result shows that, by dual-task model, 44.27\% of false positives retrieved by $DT_{embedding}$ can be successfully eliminated from the search list. Nevertheless, 10.67\% of true positives are also being erroneously pruned. Generally, for queries involving unique concepts such as \lq\lq helmet\rq\rq, \lq\lq plane\rq\rq, \lq\lq ballon\rq\rq, the accuracy of pruning is high. Query-572 \textsl{Find shots of two or more cats both visible simultaneously}, by listing \lq\lq two\rq\rq\ and \lq\lq cat\rq\rq\ as keywords, more than 60\% of false positives are removed. The method is relatively weak in interpreting queries with compositional concepts such as \lq\lq bird in the tree\rq\rq\ (query-638), only 20\% of false positives can be pruned.

Note that the setting is realistic in the scenario of interactive search. Concretely, a searcher can specify a few keywords to quickly locate a small number of true positives among the search list. These true positives identified in a short period of time can serve multiple purposes. For example, the positives are leveraged for example-based video search \cite{vitrivr_VBS2019}, online training of classifier \cite{VIREO_VBS2017} and discovery of contextually relevant concepts for query refinement \cite{VBS2018}. We repeat the experiment by randomly sampling another 30 queries from the remaining 90 queries, where the performance of $DT_{embedding}$ is generally better than $DT_{concept}$. Among the top-10 retrieved videos, there are 168 true positives and 132 false positives, implying accuracy of 56.0\%. By dual task model, 90 false positives are successfully pruned and 123 true positives are retained. The accuracy is boosted from 54\% to 74.5\%. With this, a searcher can potentially speed up the time, by skipping more than 50\% of false positives, to identify a small number of true positives for subsequent search actions.

\subsection{Boolean Queries}

Boolean expression of query terms has been exploited in manual \cite{Waseda2016} and interactive search \cite{anh:KIS}. An example of query is \textsl{Find shots of drinking beverage but not wine or beer}. The performance of concept-free search are generally unpredictable when the entire Boolean query is embedded as a single vector \cite{RUCMM2019}. A feasible way is by decomposition of query into multiple sub-queries, such as \lq\lq drinking beverage\rq\rq, \lq\lq wine\rq\rq\ and \lq\lq beer\rq\rq\ as three separate sub-queries for retrieval. The results are then merged with Boolean combination. The experiment here aims to study the limit of dual-task model in answering Boolean queries by $DT_{embedding}$ and $DT_{concept}$.

As TRECVid queries are non-Boolean, we compile 15 Boolean queries for experiment, and conduct retrieval on V3C1 dataset \cite{V3C1}. The list of queries is listed in supplementary document. For $DT_{embedding}$ and dual coding \cite{dualconding}, we contrast their performances when a query is treated as a single vector and multiple vectors by query splitting. We also compare $DT_{concept}$ with the recent work in \cite{anh:KIS} which performs Boolean based concept search using a vocabulary size of 16,263 that is similar to dual-task model. We use the measures in \cite{anh:KIS} for late fusion of search lists from multiple sub-queries for all the compared methods. As shown in Table \ref{exp:bool}, both $DT_{embedding}$ and $DT_{concept}$ generally exhibit better performances than their respective counterparts. Concept-based search by $DT_{concept}$ shows higher xinfAP not only than \cite{anh:KIS} but also concept-free search when query is not split.

An interesting note is that, when query decomposition is performed, the performance of concept-free search is boosted significantly for queries with NOT operators. Query splitting is also helpful for queries with OR operator, but is likely to hurt the performance for queries with AND operator. Nevertheless, when a query involves not only NOT but also AND or OR, answering with multiple sub-queries will bring significant improvement in retrieval. In other words, concept-free approach fails to embed the semantics of NOT in the feature space. Explicit handling of NOT operator by multiple sub-queries processing is required. The performance trend is consistent for both $DT_{embedding}$ and dual coding.

The result gives clues of how to exploit concept-free and concept-based search for Boolean queries. For example, when the query terms are logically rather than linguistically compositional, concept-based is likely a better choice than concept-free search. When NOT operator is involved, however, concept-free with query splitting strategy should be adopted. As dual-task model is interpretable with the result of $DT_{embedding}$ being verified by $DT_{concept}$ (as in Section \ref{interpretable}), appropriate search strategies can be flexibly implemented. For example, the result of concept-free search can be pruned by manually expressing the required concepts of a query with boolean expression. In general, these strategies are expected to be practical for interactive search.

\section{Conclusion}
\label{conclu}
We have presented a dual-task model to enable interpretability of concept-free (or embedding-only) search. The proposed class-sensitive BCE loss is essentially critical in guaranteeing the proper decoding of concepts. Empirical studies verify the merit of this loss function in mutually boosting the performances of both concept-free and concept-based searches under the dual-task learning strategy. A significant margin of improvement is attained when lately fusing both search lists, leading to the new state-of-the-art retrieval performances on the TRECVid datasets.

The experimental results reveal three main findings towards answering some research questions in AVS. First, the complement in search comes from different abilities in modeling the complexity of semantics. Concept-based search is limited by its ability to interpret compositional semantics that is built up from phrasal or sentence meaning. Concept-free search, on the other hand, appears less sensitive to lexical semantics that deals with individual word meaning. Late fusion of both search paradigms can lead to considerable boost in performance. Second, as proven in the empirical studies, using decoded concepts to interpret the result of concept-free search is potentially an effective strategy to eliminate false positives that rank high in the list. Such functionality will be highly practical for interactive search, where user can rapidly make sense of search result for query refinement. The final remark is that both search paradigms react differently to Boolean queries, and there is no clear way of fusion, such as by late fusion, to take advantage of each other. NOT statement, especially, remains highly difficult to be embedded and interpreted. Explicit query splitting is required to answer such queries.

\begin{acks}
This work was supported by the National Natural Science Foundation of China (No. 61872256)
\end{acks}

\bibliographystyle{ACM-Reference-Format}
\bibliography{AVS}

\end{document}